\documentclass[sigconf]{acmart}
\usepackage{enumitem}
\usepackage{balance}
\usepackage{booktabs}

\renewcommand{\paragraph}[1]{\ \\[-3mm]\noindent\textbf{#1}}

\copyrightyear{2024}
\acmYear{2024}
\setcopyright{acmlicensed}
\acmConference[SIGIR '24] {Proceedings of the 47th International ACM SIGIR Conference on Research and Development in Information Retrieval}{July 14--18, 2024}{Washington, DC, USA.}
\acmBooktitle{Proceedings of the 47th International ACM SIGIR Conference on Research and Development in Information Retrieval (SIGIR '24), July 14--18, 2024, Washington, DC, USA}
\acmISBN{979-8-4007-0431-4/24/07}
\acmDOI{10.1145/XXXXXX.XXXXXX}

\begin{document}
\title[YAGO~4.5: A Large and Clean Knowledge Base]{YAGO~4.5: A Large and Clean Knowledge Base\\with a Rich Taxonomy}
\author{Fabian M. Suchanek}
\orcid{0000-0001-7189-2796}
\affiliation{%
   \department{Telecom Paris}
  \institution{Institut Polytechnique de Paris}
 \city{Palaiseau}
  \country{France}
}
\email{fabian.suchanek@telecom-paris.fr}
\author{Mehwish Alam}
\orcid{0000-0002-7867-6612}
\affiliation{%
  \department{Telecom Paris}
  \institution{Institut Polytechnique de Paris}
  \city{Palaiseau}
  \country{France}
}
\email{mehwish.alam@telecom-paris.fr}
\author{Thomas Bonald}
\orcid{0000-0003-0468-0384}
\affiliation{%
  \department{Telecom Paris}
  \institution{Institut Polytechnique de Paris}
  \city{Palaiseau}
  \country{France}
}
\email{thomas.bonald@telecom-paris.fr}
\author{Lihu Chen}
\orcid{0009-0002-4530-7594}
\affiliation{%
  \institution{INRIA Saclay, France}
  \city{Palaiseau}
  \country{France}
}
\email{lihu.chen@inria.fr}
\author{Pierre-Henri Paris}
\orcid{0000-0002-9665-1187}
\affiliation{%
  \department{Telecom Paris}
  \institution{Institut Polytechnique de Paris}
  \city{Palaiseau}
  \country{France}
}
\email{pierre-henri.paris@telecom-paris.fr}
\author{Jules Soria}
\orcid{0009-0009-6730-7406}
\affiliation{%
  \department{Telecom Paris}
  \institution{Institut Polytechnique de Paris}
  \city{Palaiseau}
  \country{France}
}
\email{jules.soria@alumni.ip-paris.fr}


\begin{abstract} 
Knowledge Bases (KBs) find applications in many knowledge-intensive tasks and, most notably, in information retrieval. Wikidata is one of the largest public general-purpose KBs. Yet, its collaborative nature has led to a convoluted schema and taxonomy. The YAGO~4 KB cleaned up the taxonomy by incorporating the ontology of Schema.org, resulting in a cleaner structure amenable to automated reasoning. However, it also cut away large parts of the Wikidata taxonomy, which is essential for information retrieval. In this paper, we extend YAGO~4 with a large part of the Wikidata taxonomy -- while respecting logical constraints and the distinction between classes and instances. This yields YAGO~4.5, a new, logically consistent version of YAGO that adds a rich layer of informative classes. An intrinsic and an extrinsic evaluation show the value of the new resource.

\end{abstract}
\maketitle 

\section{Introduction}


A Knowledge Base (KB), also called Knowledge Graph, is a directed labeled multi-graph, where the nodes are entities (such as the United States of America, Eleanor Roosevelt, or the UN Declaration of Human Rights), and the edges are relations between these entities (such as which person is a citizen of which country, or which person contributed to which artifact)~\cite{knowledge-representation}. Similar entities are grouped into classes (such as the class of countries, the class of people, or the class of artifacts), and these classes form a taxonomy, where more general classes (such as living beings) subsume more special classes (such as humans). KBs find applications in question answering, natural language processing~\cite{machine-knowledge-survey}, and knowledge injection into language models~\cite{plugins,liu2022relational}. They are used in particular also in information retrieval, e.g., to enhance the understanding of queries and documents~\cite{reinanda2020knowledge}, to expand queries~\cite{DBLP:conf/sigir/PascaA09, DBLP:conf/sigir/DemidovaZN13}, to summarize documents~\cite{baralis2013multi}, to facilitate semantic search~\cite{ercan2019retrieving, corcoglioniti2016knowledge}, or for entity retrieval~\cite{DBLP:conf/sigir/RaeMPB12,DBLP:conf/sigir/CinarA23,DBLP:conf/sigir/JatowtKT17}.
Major industry players such as Google, Apple, Microsoft, and Meta all build and use KBs~\cite{noy2019industry}. There are also numerous public KBs, including both domain-specific ones and general-purpose ones.

\paragraph{Wikidata.} One of the largest general-purpose KBs nowadays is Wikidata~\cite{wikidata}. It provides a wealth of facts about nearly every domain of common human discourse, with more than 100 million entities and around 1.4 billion facts about them. Each entity has an abstract identifier (such as \emph{Q83396}), which makes the identifiers language-independent and persistent in time. Tens of thousands of people contribute to the project. At the same time, being a collaborative KB, Wikidata suffers from a lack of agreement on the schema level: there are several classes that are difficult to distinguish for the uninitiated user (e.g., {\it geographical location} (Q2221906), {\it location} (Q115095765), {\it geographic region} (Q82794), \emph{physical location} (Q17334923), and {\it geographical area} (Q3622002)); there are more than ten thousand relations; constraints are defined but not enforced (\emph{Grotesco} (Q10509019) is a subclass of Q49094906, which is not a class); classes and instances are mixed (\emph{scientist} (Q901), e.g., is both a subclass of \emph{person} (Q215627) and an instance of \emph{profession} (Q28640));  there are more than 2.7M classes of which only 3\% are instantiated (1M subclasses of \emph{chemical entity} (Q43460564) have no instance); and the taxonomy contains cycles (there are 47 pairs of classes that are subclasses of each other, e.g., \emph{method} (Q1799072) and \emph{technique} (Q2695280), and 15 cycles of length  3 or more, e.g., \emph{axiom} (Q17736), \emph{first principle} (Q536351), \emph{principle} (Q211364)). 
Finally, the abstract identifiers for Wikidata properties and entities make downstream applications more difficult. 

\paragraph{YAGO~4.} The YAGO KB has been in existence since 2008~\cite{yago,yago2,yago2s,yago3}. Its fourth version~\cite{yago4} was designed to address the shortcomings of Wikidata: It combines the data about instances from Wikidata with the taxonomy and properties from Schema.org -- an ontology developed by a W3C Community Group\footnote{\url{https://www.w3.org/community/schemaorg/}}.  Filtering and constraint enforcement made YAGO~4 a KB that allows for automated reasoning. However, this merger came at the expense of abandoning nearly the entire class taxonomy of Wikidata. That is a disadvantage because classes can express facts that are very hard to model correctly by RDF properties~\cite{nonnamed} -- like saying that something is a ``train ferry route'', a ``financial regulatory agency'', or a ``de facto consulate''. As a consequence, one of the major criticisms that users advanced was that the class hierarchy of YAGO~4 was too sparse.

\paragraph{Contributions.} In this paper, we show how this shortcoming of YAGO~4 can be resolved, while still maintaining the logical consistency and semantic coherence of YAGO. We carefully incorporate selected parts of the Wikidata taxonomy into the taxonomy of Schema.org. This leads to considerable challenges. Numerous organically grown branches of the Wikidata taxonomy have to be disentangled. Furthermore, many classes in Wikidata are both instances and classes and pose a challenge for modeling. 
The transformation also poses engineering challenges: Wikidata comprises more than 120 GB, even compressed, which must be parsed and processed. We will describe how we surmounted these challenges, and what open problems still remain.
The resulting resource, which we call YAGO~4.5, contains 132M facts and is logically consistent. 


Our paper is structured as follows: Section~\ref{sec:rel} recalls related work, Section~\ref{sec:design} discusses design decisions, Section~\ref{sec:implementation} discusses implementation issues, and Section~\ref{sec:result} presents the resulting KB, before Section~\ref{sec:conclusion} concludes.

\section{Related Work}\label{sec:rel}

\paragraph{General-Purpose Knowledge Bases.}
The Semantic Web comprises hundreds of KBs\footnote{\url{http://cas.lod-cloud.net/}}. Many of these are tailored for specific domains or applications, such as the Gene Ontology~\cite{geneontology} for biological processes, functions, and cellular components. However, in this paper, we are concerned with general-purpose KBs that do not focus on a specific domain. In the following, we describe several prominent general-purpose KBs. \textbf{ConceptNet}~\cite{conceptnet} is a semantic network that primarily deals with common sense knowledge. This makes it an orthogonal project to YAGO concerned with facts about instances concerning nearly all human knowledge. \textbf{BabelNet}~\cite{NavigliP12} is a large multilingual encyclopedic dictionary derived from WordNet, Wikipedia, Wikidata, and several other sources. BabelNet focuses on relations between concepts and words and has neither a taxonomy nor a schema.
\textbf{DBpedia}~\cite{dbpedia} is a large-scale, multilingual KB derived from Wikipedia (and, more recently, Wikidata). Unlike YAGO, it lacks information about the temporal validity of the facts. 
Moreover, the automated generation from the Wikipedia infoboxes and the priority for recall over precision has led DBpedia to be not fully consistent~\cite{dbpedia_inconsistency,abian2018wikidata,farber2018linked}. The manually curated part of DBpedia contains just 4M instances\footnote{\url{https://www.dbpedia.org/resources/ontology/}}. 
\textbf{Freebase}~\cite{freebase} was an extensive KB consisting of metadata compiled from various sources. The project was discontinued in 2015, and its content was transferred~\cite{pellissier2016freebase} to Wikidata. \textbf{Wikidata}~\cite{wikidata}, a project of the Wikimedia Foundation, is a collaboratively edited KB that 
supports other Wikimedia projects like Wikipedia and Wikimedia Commons. It is by far the largest open KB. However, due to its collaborative nature, its taxonomy has grown convoluted and complicated, with inconsistencies in hierarchies and data models as well as rule violations -- making it hard to use even by contributors~\cite{brasileiroApplyingMultiLevelModeling2016,piscopoWhoModelsWorld2018,shenoyStudyQualityWikidata2022}.

In this landscape, \textbf{YAGO} positions itself as a large general KB for facts about instances, with a taxonomy, manually defined properties, and logical constraints. Its key property is that it is a centrally controlled data source, which allows it to establish certain guarantees for the quality of its data~\cite{yago,yago2,yago2s,yago3,yago4}. The latest version, YAGO~4~\cite{yago4}, was designed to be clean enough to perform automated reasoning on it. However, its taxonomy is very parsimonious, which is the challenge that we address in the present paper.

\paragraph{Upper Ontologies.} Top ontologies, also known as upper or foundational ontologies, provide a domain-independent framework for organizing knowledge across various fields. We discuss some of the most prominent projects below and refer the reader to Mascardi et al.~\cite{mascardiComparisonUpperOntologies2007} for a comprehensive comparison. \textbf{Cyc}~\cite{cyc} is one of the oldest and most comprehensive upper ontologies, developed by Cycorp, aiming to represent general human knowledge and common sense reasoning. \textbf{SUMO}~\cite{sumo} is an open-source upper ontology with a formal structure for organizing and integrating domain-specific ontologies. It consists of a core set of general concepts and relations and domain-specific extensions that cover various fields, such as biology, finance, and geography.  \textbf{DOLCE}~\cite{dolce} is another top ontology 
which focuses on capturing the ontological categories underlying natural language and human cognition. \textbf{BFO}~\cite{arp2015building} is an upper ontology that was created based on the ontologies related to the domain of geospatial information. \textbf{WordNet}~\cite{wordnet} contains lexical and semantic relationships between sets of synonymous words (synsets). WordNet does not define properties, and the project was discontinued in 2012.

\textbf{Schema.org}~\cite{schema_org} differs from the above in that it is a collaborative project initiated by major companies like Google, Microsoft, and Yahoo to provide a shared vocabulary for annotating Web content with structured data. While not a top ontology in the traditional sense, Schema.org plays a crucial role in the Semantic Web ecosystem by promoting standardized vocabularies for describing entities and their properties, thus facilitating data interoperability and integration. It is broadly adopted (with more than 12 million websites in 2016) and benefits from strong industry support, making it a highly reliable and sustainable choice for building a KB~\cite{schema_stats}. 

\paragraph{Taxonomy Induction and Expansion.} Many recent studies automate taxonomy induction and expansion, including
Online Catalog Taxonomy EnrichmenT (OCTET)~\cite{kdd/MaoZKZDF020}, TaxoCom~\cite{www/LeeSKY0Y22}, TaxoOrder\cite{corr/abs-2104-03682}, TaxoExpan \cite{www/ShenSX0W020}, HiExpan \cite{kdd/ShenWLZRVS018}, TaxoEnrich~\cite{www/JiangSZ022}, and taxonomy induction from a set of terms~\cite{acl/HanRSMG18}. 
Our own previous YAGO versions automatically mapped WordNet synsets to Wikipedia categories~\cite{yago}. In contrast to these automated approaches, YAGO 4 and the new YAGO~4.5 use a manual mapping. This is because there are only a few dozen classes to be mapped. 


\paragraph{Wikidata Efforts.}
There is a community effort to map Wikidata properties and classes to Schema.org\footnote{\url{https://github.com/okfn-brasil/schemaOrg-Wikidata-Map}}. These mappings use \emph{exact match (P2888)} (64 mappings), \emph{equivalent class (P1709)} (332 mappings), \emph{equivalent property (P1628)} (84 mappings), \emph{external subproperty (P2236)} (22 mappings), and \emph{external superproperty (P2235)} (6 mappings). However, the effort was discontinued in 2017. These mappings inspired the mappings of YAGO~4, which in turn were the basis for the mappings that we use in this paper. 

Beyond that, the Wikidata community has conducted a survey on ontology issues that its contributors face\footnote{\url{https://www.wikidata.org/wiki/Wikidata_talk:Ontology_issues_prioritization\#Overview_of_potential_solutions}}. It lists in particular a ``Messy upper-level ontology'', a ``Mix-up of meta levels'', ``Exchanged sub-/superclasses'', ``Redundant classification'', ``Cycles'' (in the taxonomy), and ``Unclassified items''. However, as a community effort, Wikidata has to count on the collaboration of its contributors to solve these issues. This is a lengthy and incremental process that depends on individual commitment and consensus. The discussion dryly notes, for example, that the ``messy upper-level ontology'' is ``not fully solvable without some dictator to decide and enforce it''. This dictator, of course, cannot and should not exist in a community-driven effort such as Wikidata. In YAGO, in contrast, decisions can be taken and enforced effectively, as the team of contributors is much smaller. Indeed, YAGO has always had the strategy of ingesting instance data from large resources in a bottom-up fashion, but enforcing a top-level taxonomy, relations, and constraints in a top-down fashion. This is also the strategy that we use in this paper to extract a clean subset of the Wikidata taxonomy for YAGO.

\section{Designing YAGO}\label{sec:design}

\subsection{Design Rationale}

Our goal is to have a clean upper taxonomy for YAGO, which is precise and non-redundant to allow for automated reasoning. Our choice falls on Schema.org, for reasons we have elaborated in Section~\ref{sec:rel}: the taxonomy is concise, maintained by a W3C consortium, and finds applications well beyond its original purpose of annotating Web pages. It has the right level of detail for our purposes and does not digress into philosophical concepts. It defines not just classes, but also relations. 

One could argue that the upper-level taxonomy is sufficient and that one should not aim to add more fine-grained classes -- least of all the Wikidata taxonomy: it contains overly specific classes with few instances, some of its classes are not useful for large KB applications (such as \emph{multi-organism process} (Q22269697) for elections), and sibling classes could be further grouped into common superclasses (e.g., Wikidata is missing a class that regroups human-made places, making do instead with \emph{human-made geographic feature} (Q811430), \emph{human-made geographic feature} (Q811463), and \emph{human-made geographic object} (Q35145743)). All of this, one could argue, makes it an ill-suited candidate for a taxonomy. 

However, these issues fade when a clean upper-level taxonomy is put on top: less useful lower-level classes (such as Q22269697) disappear because they are not subclasses of the clean upper-level classes. The missing grouping, likewise, can be achieved by the upper-level taxonomy -- for example for places. Finally, even if a Wikidata class contains few instances, it can still carry meaningful information. For example, it is informative for the human user to know that an entity is a ``General aviation monoplane with 1 tractor-piston-propeller engine'' (Q33110974). It was precisely that lack of such classes that users deplored for YAGO~4. Such classes do not carry logically formalized meanings. (It would be cumbersome to formally express that something is a ``General aviation monoplane with 1 tractor-piston-propeller engine'' by RDF statements~\cite{vagueness,nonnamed}.)
Rather, the purpose of these lower-level classes is to convey informal information to the human user. The Wikidata classes clearly serve this purpose.\\[-5mm]

\subsection{Design Principles}\label{sec:designprinciples}

Our goal is to integrate the upper taxonomy from Schema.org with the lower taxonomy from Wikidata. The following design principles drive our integration: 

\paragraph{1. Prefer properties over class membership.}  Some information can be expressed either by a property (\emph{hasNationality United\-States}) or by a class membership (\emph{type American}). When designing the schema, we give preference to properties and choose classes only if the class appears in the domain or range of a property. In our example, \emph{American} does not appear as a range or domain of any property (its superclass \emph{Human} does). Hence, the class \emph{American} should not exist, and the nationality should be expressed by a property. The reason for avoiding classes when possible is that OWL DL (the reasoning formalism we target) does not allow expressing properties about classes (i.e., we cannot attach properties to the class \emph{American}, while we can for the instance \emph{United\-States}). This choice is also consistent with Wikidata's and YAGO's way of modeling. 

\paragraph{2. Choose the property with fewer objects.} When we have the choice between a property and its inverse property (e.g., \emph{hasCitizenship} and \emph{hasCitizen}), we choose the one that has, on average, fewer objects per subject (i.e., \emph{hasCitizenship}). The reason for this choice is that it allows seeing the properties as attributes ``about'' the subject. For example, the Wikipedia page of Eleanor Roosevelt lists her US-American nationality, because the nationality is perceived as a property ``of a person''. At the same time, the Wikipedia page about the United States does not list all people of American nationality, because a citizen is not perceived as a property ``of a country''. Our criterion formalizes this intuition. It is indeed \textit{de facto} used by all major KBs~\cite{amieplus}.

\paragraph{3. The upper taxonomy exists to define formal properties that will be populated.} All classes of the upper taxonomy shall define formal properties (e.g., an \emph{Airline} has an \emph{iata\-Code}, which justifies its existence as an upper-level class). Both the domain and the range of these properties have to be upper-level classes. The reason for this design choice is that it makes the upper taxonomy a self-contained schema. Schema reasoning can then be restricted to this upper-level taxonomy without the need to search for property definitions in the lower classes.  

\paragraph{4. The lower taxonomy exists to convey human-intelligible information about its instances in a non-redundant form.} While the upper-level taxonomy contains the schema, the lower-level taxonomy targets mainly human users. Our design principle thus tells us to remove classes that add no information over upper taxonomy classes, to eliminate links in the taxonomy that are redundant due to transitivity, to merge classes that are hard to distinguish for the uninitiated user, and to remove classes that are not populated.

\subsection{Upper Taxonomy}\label{sec:top-level}

We now discuss how we construct the upper-level taxonomy of YAGO~4.5.

\paragraph{Upper-level classes.} As for YAGO~4~\cite{yago4}, we start with the taxonomy of Schema.org. It defines one top-level class \emph{schema:Thing} with 11 subclasses. We exclude the class \emph{Action}\footnote{For ease of reading, we omit prefixes where these can be inferred.}, which models mainly Web user actions. We exclude \emph{Bio\-Chem\-Entity} and \emph{Medical\-Entity} because these are domain-specific concepts. This leaves us with 8 top-level classes (\emph{CreativeWork, Event, Organization, Taxon, Person, Place, Product, Intangible}), which we accept as subclasses of \emph{Thing}. All top-level classes are declared disjoint (except places/organizations, and products/creative works). 

\paragraph{Fictional entities.} To deal with fictional entities, we add a class \emph{yago:FictionalEntity} as a subclass of \emph{schema:Thing}. This class defines properties such as \emph{yago:createdBy} and \emph{yago:appearsIn}, thus justifying its existence as a class under Design Principle 1. Fictional entities are not disjoint with any other class, as anything can also exist in fiction. A fictional entity is an instance of both \emph{yago:FictionalEntity} and the class it belongs to in fiction. For example, a fictional human is an instance of both \emph{yago:FictionalEntity} and \emph{schema:Person}. This has the disadvantage that fictional humans will be counted as humans in count queries. However, it has the advantage that one can easily reason on fictional humans, as they will share all the properties that we declared for \emph{schema:Person}. Wikidata goes a different way, by recreating the entire class hierarchy with its properties also for fictional beings, mapping each class of fictional entities to its real-world class counterpart. This choice can for sure be defended, but for YAGO, it would have severely convoluted the schema: it would have required duplication of all class and property specifications. The current modelization already has an advantage over the modelization in previous versions of YAGO (which simply merged real and fictional entities), as well as over other top-level ontologies such as DOLCE, BFO, and SUMO (which do not model fictional entities at all). 

\paragraph{Intangibles.} For our new YAGO, we added the following classes that are not in Schema.org, but that are necessary to define the ranges of properties (under Design Principle 3 above): \emph{yago:Award}, \emph{yago:Gender} (which differs from \emph{schema:GenderType} in that it allows more than two values), and \emph{yago:BeliefSystem} (for religious adherence). All are subclasses of \emph{schema:Intangible}. The other subclasses of \emph{Intangible} that Schema.org defines are mostly Web-specific (e.g., \emph{Action\-Access\-Specification}). Since these classes would not have instances, let alone populated properties from Wikidata, we removed them under Design Principle 3. 

Schema.org has a subclass \emph{Occupation} of \emph{Intangible}, and models occupations by the property \emph{has\-Occupation}. However, if we model occupations by a property, (1) we lose the class hierarchy of occupations (\emph{Physicist subClassOf Scientist} etc.), and (2) we lose the ability to add properties to specific professions (such as the \emph{doctoral advisor} for scientists). Hence, by Design Principle 1 above, we model professions rather as subclasses of \emph{Person}.

\paragraph{Places.} When it comes to places, the taxonomy of Schema.org is heavily oriented towards the annotation of Web pages, with subclasses such as \emph{Accommodation}, \emph{Residence}, \emph{Local\-Business}, etc. These classes do not define properties that we could populate from Wikidata, and therefore, we remove them under Design Principle 3. We then manually created a taxonomy of subclasses of \emph{schema:Place}, which distinguishes \emph{schema:Landform} (areas with a boundary given by nature), \emph{schema:Administrative\-Area} (boundary given by human administration) and the newly created \emph{yago:Human\-Made\-Place} (boundaries given by human physical construction) and \emph{yago:As\-tro\-no\-mical\-Object} (with boundaries in space). For the former, we add a subclass \emph{yago:Way}, which regroups all ways of transit (roads, canals, railway lines, etc.). 

\paragraph{General considerations.}
Under Design Principle 3, we keep only those classes from Schema.org that add new properties (plus their super-classes, all the way up to \emph{schema:Thing}). This results in 41 upper classes. As in YAGO~4, all of the above is expressed as SHACL constraints\footnote{\url{https://www.w3.org/TR/shacl/}} on the classes of Schema.org. Thus, there is no special syntax, code, or formalism for these declarations, and they are all part of the YAGO KB as normal facts.

\subsection{Lower Taxonomy}\label{sec:wikidatatax}

\paragraph{Mapping to Wikidata.} The lower levels of the YAGO~4.5 taxonomy come from Wikidata. As in YAGO~4, each class in the YAGO upper taxonomy is manually mapped to one or more classes in Wikidata. This happens, likewise, in a fully declarative way with a simple RDF statement that links the Schema.org-class by a special predicate to the Wikidata class(es). The mapping can happen at any level of the upper taxonomy: general classes such as \emph{Organization} are mapped to Wikidata, and more special classes such as \emph{Corporation} are mapped as well. A mapping can give rise to the following constellations:

\begin{description}[leftmargin=5mm]
    \item[One-to-one mapping.] One upper class is mapped to one Wikidata class. All of the subclasses of the Wikidata class are glued under the upper class, but its super-classes are not imported.
    \item[One-to-many mapping.] One upper class is mapped to several Wikidata classes. Again, all subclasses of these are glued under the upper class. This has the effect of merging the Wikidata classes. We do this, e.g., for classes that are equivalent for our purposes (such as \emph{geographical region} (Q82794) and \emph{geographical area} (Q3622002)).
    \item[One-to-none mapping.] One upper class is mapped to no Wikidata class -- only its subclasses are mapped to Wikidata. This has the effect that there cannot be direct instances of the upper class. Nor can there be subclasses other than the ones we declared. This is a new mechanism that did not exist in YAGO 4. We use it for classes with a convoluted taxonomy in Wikidata.
\end{description}

\noindent We use a one-to-none-mapping for the following classes:
\begin{description}[leftmargin=5mm]
    \item[schema:Thing.] In YAGO~4, this class was mapped to the top-level class \emph{entity} (Q35120) in Wikidata, which resulted in more than 1 million direct instances of schema:Thing in Yago. This defies Design Principle 3, because \emph{schema:Thing} defines only very few properties. Hence, we now accept only entities that fall into one of the manually approved subclasses of \emph{Thing}. This results in a clean top-level taxonomy, discarding meta-classes (e.g., \emph{class or metaclass of Wikidata ontology} (Q21522864)), overlapping classes (e.g., \emph{geographic entity} (Q27096213) and \emph{location} (Q115095765))
        and too specific classes (there are 6 top classes in Wikidata with less than 40 instances each, e.g., \emph{converter} (Q35825432)). 
    \item[schema:Place.] The taxonomy of Wikidata for places is highly convoluted, with classes that are difficult to distinguish such as \emph{terrain} (Q14524493), \emph{geographical location} (Q2221906), \emph{geographical region} (Q82794), \emph{geographical area} (Q3622002), and \emph{location} (Q115095765).     Hence, we do not map \emph{schema:Place}, and accept only instances of its manually designed subclasses. This discards 2,861 classes (from 29,826, i.e., less than 1\%) and 137k instances (from 19M, i.e., less than 1\%).
   \item[schema:Intangible.] Intangible classes in Wikidata, likewise, are highly convoluted, with classes such as \emph{class} (Q5127848), \emph{process} (Q3249551 and Q67518233)  or \emph{role} (Q4897819). Hence, here, too, we would have difficulties establishing properties to comply with Design Principle 3. Therefore, we apply a one-to-none mapping and accept only instances and subclasses of the manually declared subclasses of \emph{Intangible}.
    \end{description}

\paragraph{Importing the subclasses.} Once the mapping has been defined manually, the subclasses of Wikidata can be imported automatically. To this end, we consider the subclass graph of Wikidata, which we construct as follows: every entity of Wikidata that has a \emph{subClassOf} relationship becomes a node in this graph. It is linked to its superclasses by the \emph{subClassOf} relationship. We then iterate through all upper classes of YAGO, and whenever we hit a class that is mapped to Wikidata, we glue the entire sub-DAG of that Wikidata class under the YAGO top-level class. This approach differs from the approach in YAGO~4, where we mapped only classes with a Wikipedia article. However, this restriction limited the taxonomy to just a handful of classes per instance, which proved to be too few.

Several caveats are to be respected in this merging process to respect Design Principle 4: as in YAGO~4, we ensure that we do not add any link that would create a loop in the taxonomy (57 loops were removed). We do not add transitive links (40k such links were removed). We do not add a link if that would make a class a transitive subclass of two top-level classes that we declared disjoint (9k links were removed). We also remove, from the sub-DAG that we import, all Wikidata classes that have their own mapping to YAGO upper classes. 

\paragraph{Excluded classes.} We exclude housekeeping classes of Wikidata that we blacklist manually: classes of Wikimedia pages, disambiguation pages, lists, and the like. For our new YAGO, we also exclude linguistic objects (such as characters (Q3241972), phrases (Q187931), numbers (Q11563), etc.), many of which are technically infinite and would otherwise make up 700k entities in YAGO. We also remove abstract objects such as actions and occurrents, as these have rather philosophical subclasses that are of limited use for our purposes (e.g., \emph{Multi-organism process} (Q22269697), etc.). The same fate is bestowed on scholarly articles. The addition of all obtainable scholarly articles to Wikidata was controversial\footnote{\url{https://www.mail-archive.com/wikidata@lists.wikimedia.org/msg06716.html}}, and in YAGO, they would make up 39M of entities, almost half of all entities. Hence, we decided to remove them. Finally, under Design Principle~4, we also remove all classes that do not have instances (1.3M).

\subsection{Instances}\label{sec:instances}

\paragraph{Identifiers.} As in YAGO~4, every instance is automatically equipped with a readable name. We use the title of the corresponding Wikipedia page as an entity identifier (as in \emph{yago:Eleanor\_Roosevelt}). If there is none, or if the same Wikipedia page is used by more than one entity, we use the English label of the entity and concatenate it with the Wikidata Q-id to avoid ambiguity (as in \emph{yago:\-Brazilian\_\-jiu\_\-jitsu\_\-competition\_\-Q105086361}). If there is none, we use a label that contains legal Turtle characters, concatenated with the Wikidata id (which was not done in YAGO 4). If there is no such label, we use the Wikidata id (with a YAGO-prefix for uniformity). The Turtle standard\footnote{\url{https://www.w3.org/TR/turtle/}} allows percentage codes in local names, but many parsers (e.g., the one of Hermit\footnote{\url{http://www.hermit-reasoner.com/}}) cannot deal with them. Hence, we replace all characters that are not letters or numbers by their hexadecimal Unicode, so that two identifiers that differ only in their inadmissible characters are still distinct. 

\paragraph{Instances vs. Classes.} Wikidata contains several items that are both instances and classes. For example, \emph{English} (Q1860) is an instance of \emph{Natural Language} (Q33742), as well as a subclass of \emph{Anglic} (Q1346342). That makes sense because there can be several subclasses of \emph{English}, such as e.g., \emph{American English} (Q7976). As per our discussion in Section~\ref{sec:wikidatatax}, \emph{English} thus becomes a class. At the same time, we would also like to say that Eleanor Roosevelt spoke English, i.e., we would like to make a statement about a class. The dominant reasoning language, OWL 2, allows such statements by a mechanism called \emph{punning}\footnote{\url{https://www.w3.org/TR/owl2-new-features/\#F12:\_Punning}}. However, OWL 2 punning works essentially by allowing the same identifier (\emph{English}) to denote two distinct elements (a class and an instance). This means that neither standard description logic reasoners nor the query language SPARQL can deal with expressions where a variable is bound to an identifier that is simultaneously a class and an instance~\cite{LENZERINI2021103432}.
Thus, it would be impossible to say that Eleanor Roosevelt spoke a subclass of Anglic. (This problem did not appear in YAGO~4, where intermediate classes were eliminated aggressively, and survived only as instances.) We solve this problem as follows: whenever we encounter a Wikidata fact that would link an instance to a class, we create a generic instance of the class and use it as an object. In our example, we create a fact saying that Eleanor Roosevelt spoke \emph{yago:English\_language\_generic\_instance}. The intuition is that Roosevelt spoke something that is an instance of English. This mechanism is in line with Approach~2 in \cite{prop_values}, and kicks in for awards, belief systems, academic titles, and languages.  

This technique works well for the objects of statements, which generally have an existential interpretation (Eleanor Roosevelt did not speak all dialects of English, but there is one that she spoke). It works less well for subjects of statements. For example, commercial products are classes in Wikidata (and rightly so, as different people can own different instances of the same product). However, if we take commercial products as classes, we cannot attach their manufacturer, date of inception, awards, etc. to them. It would be semantically wrong to attach these to a generic instance of the product, as they apply to the line of products itself. It would also be semantically wrong to create an axiom that says that all instances of the class have that property (not every single iPhone has won an award). Hence, we make every item that is a product (as identified by the \emph{manufacturer} (P176) relation in Wikidata) an instance. 

\subsection{Properties and Constraints}\label{sec:props}

Properties and constraints work largely as in YAGO~4: for \emph{schema:Thing} and each sub-class from Schema.org, we manually define the properties that Schema.org provides. By Design Principle~3, we keep only those properties of Schema.org that are (1) of general interest (removing specialized properties such as \emph{has\-Drive\-Through\-Service}) and (2) existent in Wikidata (because otherwise they would not be populated). For the new YAGO, we added a few properties, most notably for the new top-level classes \emph{yago:Award} and \emph{yago:FictionalEntity}, and also for countries (Schema.org does not contain the relation \emph{has\-Capital}). Each of the 100 most frequent relations in Wikidata is either mapped to YAGO or excluded on purpose (e.g., because of redundancy). Our design document\footnote{\url{https://yago-knowledge.org/data/yago4.5/design-document.pdf}} lists the most frequent Wikipedia properties, together with their mappings to YAGO~4.5.

An instance can have only those properties that are declared for its class or superclasses. We manually add SHACL constraints~\cite{shacl} for maximum cardinality (31 constraints in total), patterns of literals (e.g., for ISBNs; 9 in total), and domain and range constraints (for each relation). Each relation is mapped manually and declaratively to a relation in Wikidata, and populated from there. A fact is accepted only if both its subject and its object conform to the domain and range constraints (this removes roughly 6\% of facts in our dataset). We take only the facts that Wikidata labels as ``truthy'' (which exclude disputed statements). We extract time stamps for facts from Wikidata, and attach them to the YAGO facts in the RDF-star model~\cite{rdf-star}.

The entire taxonomy of upper classes, the definition of properties, the accompanying SHACL constraints, and the mapping to Wikidata classes take the form of a single Turtle file. It forms an integral part of YAGO and can be downloaded along with the rest of the KB. All of this works exactly as in YAGO~4, and we refer the interested reader to the corresponding publication~\cite{yago4}.

Unlike YAGO~4, the new YAGO simplifies numerical quantities: while these were previously values with a range and a unit, they are now simple literals. Our design document$^{12}$ lists, explains and justifies the design decisions for each class and relation in detail.

\section{Implementing YAGO}\label{sec:implementation}

Parsing, analyzing, and transforming a KB of the size of Wikidata (766 GB as of April 2023) is no easy feat. We describe here the challenges we encountered when creating YAGO~4.5, and how we surmounted them, in the hope that this will be useful for other users of Wikidata and YAGO.

\paragraph{Infrastructure.} The code of YAGO~4 was written in Rust. While this ensured high performance and compile-time flagging of code problems, it also complicated the maintenance of the project
, and we have, therefore, rewritten the code from scratch in Python. The original YAGO~4 loaded the data into a RocksDB key-value store. This had the advantage that many costly operations (such as constraint checks) could be run directly on the data store. At the same time, loading the entire data into the data store and indexing it could easily take a day. The new system, therefore, stores all data (intermediate and final) in files on the hard drive, which has the advantage that intermediate results can be inspected and re-used. 

\paragraph{Data formats.} Wikidata exists in the ``full'' version and the ``truthy'' version\footnote{\url{https://www.wikidata.org/wiki/Wikidata:Database_download}}. While the truthy version is much smaller, we need the full version to extract time stamps for facts (a feature YAGO has had since Version~2~\cite{yago2}). One has the choice between the NT format (easier to parse) and the Turtle file (smaller by a factor of 2), and our choice falls on the latter. The file can be downloaded as BZ2 and GZIP, and we strongly recommend the GZIP version. First, the unpacking is much faster (in the order of hours instead of the order of days in our case). Second, BZ2 allows no way of seeing the compressed file size without unpacking it. Finally, GZIP files can be processed sequentially by Python without unpacking them. For parsing Wikidata, we experimented with \href{https://pypi.org/project/rdflib/}{RDFlib}. However, the library failed for certain characters in URIs, which caused an unrecoverable abortion in the middle of the parsing. Furthermore, the generation of URIs (expanding the prefixes) causes a large overhead. Therefore, we wrote our own Turtle parser and graph database, which, for our limited application, turned out to be rather simple (500 lines of code). These design choices mean that our code does not use any external libraries. 

While our initial input (Wikidata and schema.org) is in Turtle, we chose TSV as our intermediate file format, because it allows for much faster parsing. In addition, we can attach more information to each fact (time stamp, source, etc.) in the form of supplementary columns. 

\paragraph{Data processing.} Our system proceeds in 6 sequential steps. Each step reads the output files of the previous step, and produces new output files, as in \cite{yago2s}. Each of these steps can be run on its own, and each of the steps has its own set of test input files with gold-standard output files, which we can use to check if the step works as expected.

Two of our steps need to process the entire Wikidata file, which is done in parallel. We experimented with Python multithreading, only to find that, due to the Global Interpreter Lock, it does not fully utilize multiple processor cores for CPU-bound tasks. The correct construction is multiprocessing, which uses one processor per process. Since processes cannot efficiently share data, each process has to load a copy of the data that it needs. Each process writes out its results to its own temporary file, which we then merge with the others. Each process $i$ of the $n$ processes starts at position $(i-1)/n\times{}N$ of the Wikidata file (where $N$ is the size of the file). From that position on, the process scrolls forward to the next item declaration, where it starts its work. It proceeds until it hits the item declaration that follows position $i/n\times{}N$. Since the file is UTF-8 encoded, the initial position may hit the middle of a character that is encoded by more than one byte. This is not a problem because the UTF-8 standard can distinguish the middle-bytes from the initial bytes in an encoded stream.

\paragraph{Steps.} We generate YAGO on a Unix machine with 90 CPUs and 800GB of RAM. We proceed in 6 steps:

\begin{description}[leftmargin=3mm,nolistsep]
    \item[1. Create schema:] The manual definition of the schema is loaded, and the relevant parts of the Schema.org-taxonomy are extracted, as described in Section~\ref{sec:top-level}. This process operates only on manually defined files and hence terminates on our machine in less than a second.
    \item[2. Create taxonomy:] Wikidata is parsed for classes, and a loop-free taxonomy is constructed, as described in Section~\ref{sec:wikidatatax}. This step is parallelized and takes 4 hours. 
    \item[3. Create facts:] Wikidata is parsed for facts, each predicate is mapped to a YAGO predicate as described in Section~\ref{sec:props}, the subject of the fact is type-checked, and the objects are type-checked if they are literals. The objects that are not literals cannot yet be type-checked because we do not yet have a complete list of all instances at this stage. This step, likewise, is parallelized and takes 4 hours. 
    \item[4. Type-check facts:] The previous step has given us a list of facts, which also contains the class that each instance belongs to. We load this list into memory and run through all facts to type-check the object of each fact. This step runs on YAGO data, and not on Wikidata, and thus does not need parallelization. It takes 1:30h to run.
    \item[5. Create ids:] Among those facts that survived the type check, we map each entity to its legible YAGO name, as described in Section~\ref{sec:props}. This takes one hour.
    \item[6. Create statistics:] Debugging and testing are an integral part of the development. The last step counts the number of instances per class and of facts per predicate. It creates a visualization of the taxonomy and a random selection of entities for manual check. This process also takes one hour. 
\end{description}

\noindent Thus, the overall process takes about 12 hours.

\section{Result}\label{sec:result}

\subsection{Resource}

\paragraph{Size.} Table~\ref{tab:stats} shows the statistics of YAGO~4.5\footnote{Version yago-4.5.0.2} and puts them in perspective with Wikidata and YAGO 4. Both versions of YAGO have vastly less predicates than Wikidata. As explained previously, this is deliberate: according to Design Principle 3 (Section~\ref{sec:designprinciples}), YAGO accepts only those predicates that are defined in its schema. That said, YAGO~4.5 does cover most of the 100 most frequent predicates of Wikidata (see Section~\ref{sec:props}). When we now turn to compare the two versions of YAGO, we see that YAGO~4.5 has fewer properties than YAGO~4. This is due to the removal of inverse properties, which we removed under Design Principle 2  (e.g., we removed \emph{hasParent} because we already have \emph{hasChild}; 6 such cases); properties related to scholarly articles (\emph{citation} etc., 4 in total, which we removed according to the discussion in Section~\ref{sec:wikidatatax}); biochemical properties (11 in total, removed because of a lack of expertise in our team); and properties of numerical literals (\emph{value} etc., 6 in total, removed because literals became simple values in YAGO~4.5). The loss is thus deliberate. A detailed comparison of the properties in YAGO~4, YAGO~4.5, and Wikidata is in our design document$^{11}$. The same goes for the facts of YAGO 4: 238M facts describe scholarly articles (mainly citations, pagination, and publication dates), 13M facts describe literals and 2.5M facts are redundant because of an inverse property. If these are discarded, the new YAGO contains slightly more facts. The dump size is still smaller because we use the Turtle file format instead of NT. Concerning the classes, the picture is as rosy as intended: YAGO~4.5 has vastly more classes. 

\paragraph{Data Format.} The final file format of YAGO is Turtle.
We separate the subject, predicate, object, and dot by a tabulator, so that our files are \textit{de facto} also TSV files. YAGO is split into the following files:
\begin{itemize}[leftmargin=5mm]
    \item\textbf{Schema}: upper taxonomy, property definitions, and SHACL constraints.
    \item\textbf{Taxonomy}: the entire taxonomy of YAGO (all \emph{subClassOf} facts).
    \item\textbf{Facts}: all facts about entities that have an English Wikipedia page.
    \item\textbf{Beyond Wikipedia}: all facts about other entities.
    \item\textbf{Meta}: all temporal annotations (in the RDF-star file format~\cite{rdf-star}).
\end{itemize}
Downstream applications can load only the files they need, and exclude, e.g., meta facts or facts about entities that do not have an English Wikipedia page. 

\begin{table}
    \centering
    \begin{tabular}{lrrr}
    \toprule
         & ~~~~Wikidata &  ~~~~~~~~~YAGO 4 & ~~~~YAGO~4.5 \\
         \midrule
         Entities & 103M & 67M (37M) & 49M\\
         ~~~of which generic & 0 & 0 & 62k\\
         Classes & 2.8M & 10k  & 133k \\
         Predicates & 11k & 140 (124) & 108 \\
         Facts & 500M & 343M (89M) & 132M \\
         Type facts & 106M & 70M (33M) & 53M\\
         Label facts & 795M & 303M & 479M\\
         Meta facts & 12M & 2.5M & 7M\\
         Dump size  & 766GB & 280GB & 142GB\\
         \bottomrule\\
    \end{tabular}
    \caption{Size of YAGO~4.5. Facts exclude \emph{type}, \emph{label}, \emph{comment}, \emph{alternate\-Name}, \emph{same\-As}, and \emph{main\-Entity\-Of\-Page} facts. In brackets: without redundant properties, properties describing literals, and properties describing scholarly articles. }     \label{tab:stats}
\end{table}

\subsection{Evaluation}

\begin{table*}
    \centering
    \begin{tabular}{llrrr}
    \toprule
         Criterion & Operationalization &  ~~~~Wikidata &  ~~~~YAGO 4 & ~~~~YAGO~4.5 \\
         \midrule
         Consistency & Absence of contradictions & no & \textbf{yes} & \textbf{yes} \\
         Complexity & Top-level classes & 41 & 2714 & \textbf{9} \\
                    & Number of paths to root & 44 & \textbf{1.1} & 2.3 \\   
         Modularity & Disjointness axioms & 0 & 18 & \textbf{24}\\
         Conciseness & Taxonomic loops & 62 & 0 & \textbf{0}\\                      
         & Redundant taxonomic links & 377k & 1216 & \textbf{0}\\                      
         & Redundant relations & 118 & 6 & \textbf{0}\\
                     & Classes without instances & 2.6M & 73 & \textbf{0} \\ 
         Understandability~~~ & Human-readable names & 0\% & 89\% & \textbf{91\%}\\
         Coverage    & Classes per instance & \textbf{8.4} & 3.6 & 7.8 \\    
                              & Facts per instance & 4.8 & \textbf{5.1} & 2.7\\
         \bottomrule
    \end{tabular}
    \caption{Quality measures, inspired by~\cite{ontology-evaluation}}\label{tab:quality}
\end{table*}
\paragraph{Intrinsic evaluation.} To evaluate the quality of the new KB, we draw inspiration from the criteria for ontology evaluation that a recent survey identified~\cite{ontology-evaluation}: \textbf{consistency} refers to the absence of logical contradictions. 
For this purpose, we verified the logical consistency of YAGO using the OWL API\footnote{\url{https://owlcs.github.io/owlapi/}} and Pellet~\cite{pellet}, an OWL DL reasoner for Java, in 4 hours. Additionally, we validated the SHACL shapes with Jena SHACL\footnote{\url{https://jena.apache.org/documentation/shacl/}} in 1h30.
\textbf{Complexity} refers to the extent to which the ontology is complicated. We propose to measure it (1) by the number of top-level classes (i.e., the number of direct subclasses of \emph{Thing}; the fewer the better), and (2) by the average number of paths from an instance to the root in the taxonomic tree (the fewer the better). \textbf{Modularity} is the degree to which the ontology is composed of discrete subsets. In our case, these subsets are the disjoint classes, and we report the number of (non-redundant and actively enforced) class disjointness statements as an indicator. \textbf{Conciseness} requires the absence of redundancies, and we count the number of taxonomic loops, taxonomic links that are redundant because of transitivity, and relations whose inverse also exists. \textbf{Understandability} is the degree to which the ontology can be comprehended. This is difficult to operationalize, but we can at least report the percentage of identifiers that have human-readable names. \textbf{Coverage} refers to the degree to which the ontology covers the domain knowledge. We report the number of classes and facts per instance (excluding labels etc.).
 
Table~\ref{tab:quality} shows these measures for YAGO 4, YAGO~4.5, and Wikidata. The latter has vastly more facts per instance than YAGO. This is to be expected, as YAGO is a clean subset of Wikidata. YAGO~4, too, has more facts per instance than YAGO~4.5. However, this is mainly due to the 174M facts about citations for 40M scholarly articles. YAGO 4 also has a 
lower number of paths to the root, which is due to its sparse taxonomy. On all other measures, YAGO~4.5 scores much better than both YAGO~4 and Wikidata: it has a clean upper-level taxonomy of just nine top-level classes -- instead of the dozens of Wikidata or the thousands that YAGO 4 attached to the taxonomic root for lack of good intermediate classes. YAGO~4.5 is also free of all types of redundancy that plagued Wikidata and YAGO 4. On average, 
there are just 2.6 paths to the root, as opposed to 44 in Wikidata (not accounting for cycles). At the same time, YAGO~4.5 nearly replicates the taxonomic richness of Wikidata, with 7.8 classes per entity -- twice as many as in YAGO 4.

\paragraph{Extrinsic evaluation.} To show the value of the new YAGO~4.5 over YAGO 4, we applied it to the task of entity disambiguation (also called entity linking). Given a KB and a text that contains an entity mention, the task is to link it to the corresponding entity in the KB. This task is important for analyzing large document sets such as the Panama Papers
\footnote{\url{https://medium.com/@ambiverse}} 
or newspaper archives~\cite{economist}, and it is useful also in information retrieval~\cite{zwicklbauer2016robust,entitylinkingsigir}. Entity disambiguation is particularly difficult with ambiguous entity names. For example, in ``I love the city of light, Paris'', the word ``Paris'' could refer to the capital of France, but also to several dozen cities of that name in the US, not to mention people of that name such as Paris Hilton or the Greek hero Paris. We use the BLINK~\cite{wu2020scalable} dataset, which is based on a 2019 English Wikipedia dump. We collect 19 thousand samples from this dataset in such a manner that (1) every mention appears only once (so that there is less bias towards popular entities), and (2) all entities exist in both YAGO 4 and YAGO~4.5. For each entity mention, we collect the possible candidate entities by finding all entities that share a label with the mention. We use the end-to-end entity linking system ExtEnD~\cite{barba-etal-2021-extend}, pre-trained on Longformer~\cite{beltagy2020longformer}. 
It takes as input a single piece of text, which consists of the input text concatenated with the separator [SEP] and a sequence of entity candidates. Each entity candidate is given in textual form. For our experiments, we use the main label of the entity, followed by the classes of which the entity is an instance. In our example, we produce ``I love the city of light, \textit{Paris} [SEP] Paris [capital, city, place]; Paris Hilton [singer, actress, human]; ...''. 
Then, the system indicates the start and end token of the span containing the predicted entity candidate. 

Table~\ref{tab:candidate_num} shows the accuracy of the disambiguation, grouped by the number of candidate entities per mention. The disambiguation works much better on YAGO~4.5 -- especially on mentions that have many candidates. This shows the usefulness of the new YAGO taxonomy. All code and data of this study are publicly available\footnote{\url{https://github.com/tigerchen52/eval_yago_el/}} for reproducibility.

\begin{table}[tp] 
	\centering 
	\small
			\begin{tabular}{cccccc}  
				\toprule  
				&\textbf{[1, 5)}&\textbf{[5, 10)}&\textbf{[10, 20)}&\textbf{[20, $\infty$)}&\textbf{Macro}\cr
            samples&13971&3452&1374&203&19000\cr
            \midrule
YAGO 4&77\% &59\% &53\%&20\%&52\%\cr
            YAGO~4.5 &\textbf{80\%}&\textbf{64\%}&\textbf{58\%}&\textbf{31\%}&\textbf{58\%}\cr	
            \bottomrule
			\end{tabular}
			\caption{Disambiguation accuracy by candidate  mention.} 	\label{tab:candidate_num}
\end{table}

\subsection{Applications}\label{sec:applications}

\paragraph{Benchmarking.} Previous versions of YAGO have been widely used as benchmark datasets in 
entity type prediction~\cite{HuGX0P23} and link prediction~\cite{GeseseBAS21}. YAGO43KET~\cite{moon2017learning} is a benchmark dataset for entity type prediction, which has been created from the subset of YAGO 3.0. It contains  43k entities with their type information. 
YAGO3-10 is a benchmark for link prediction that consists of all entities of YAGO 3.0 having at least 10 predicates ~\cite{dettmers2018convolutional}; it consists of 120k entities (mainly people)  and 37 predicates. YAGO11K~\cite{dasgupta2018hyte} is also a subset of Yago 3.0, and it is used to evaluate algorithms for temporal link prediction~\cite{WuXZMCC21,LiJLGGSWC21} and 
temporal relation prediction~\cite{ChenXS0D23}.

\paragraph{YAGO in Information Retrieval.}
YAGO is helpful for Information Retrieval (IR) systems in order to enhance their understanding of queries and documents beyond the scope of word tokens and plain texts~\cite{reinanda2020knowledge}.  YAGO can help IR mainly in two applications: Document Retrieval and Entity Retrieval.
\emph{Document retrieval} is the task of finding relevant textual resources for a given user query. Several works leverage entity information from YAGO to expand queries, incorporating rich features into the retrieval process~\cite{DBLP:conf/sigir/PascaA09, DBLP:conf/sigir/DemidovaZN13}.
Another approach uses YAGO ontology to construct document summarization tools, aiding in the efficient retrieval of key information from extensive document collections~\cite{baralis2013multi}.
Facts in YAGO can also be used to represent both queries and documents, facilitating semantic search in information retrieval~\cite{ercan2019retrieving, corcoglioniti2016knowledge}.
\emph{Entity Retrieval} aims to retrieve and rank entities in a document collection (or a knowledge base). Several benchmarks for entity retrieval are derived from YAGO~\cite{hoffart2011robust, noullet2020kore, chen2024learning}.
Existing studies rely on information in YAGO to retrieve various entities, including geographic entities~\cite{DBLP:conf/sigir/RaeMPB12}, Point of Interest (POI) entities~\cite{DBLP:conf/sigir/CinarA23}, and temporal scopes of entities~\cite{DBLP:conf/sigir/JatowtKT17}.


\paragraph{Other Applications.}
The YAGO KB has found quite a number of other applications in the past~\cite{yago-resource}. Together, the YAGO publications have been cited more than 10,000 times in total so far, according to Google Scholar. As every user is free to download the KB and to use it (or not), we do not have an overview of the projects that use YAGO. We just know that the KB (in various versions) has been downloaded about 6000 times during the past year (heuristically excluding bots). YAGO~4.5 can be plugged in as-is into any system that uses YAGO 4 (as is currently happening with Qlever~\cite{qlever}). We expect our YAGO~4.5 to be even more useful than YAGO 4, as it provides fine-grained information about instances by the new classes. This will be useful to any application that draws on classes: entity retrieval, semantic annotation of documents, graphical exploration of instances, similarity computations of entities, of documents, or of queries and documents, and, as shown, entity disambiguation. 

\paragraph{Availability.} YAGO~4.5 is available on the YAGO Web site\footnote{\url{https://yago-knowledge.org}}. It contains download links for the data, the Design Document that details the mapping of schema.org to Wikidata, the documentation of the data format, the list of publications, and the names of the contributors. The Web page also offers an interactive browser for the KB (loaded with a subset of the entities). A SPARQL endpoint is provided\footnote{\url{https://yago-knowledge.org/sparql}}. The URIs of YAGO~4.5 are dereferenceable. YAGO~4.5 is available with Creative Commons Attribution-ShareAlike License (as imposed by the license of schema.org). The source code of YAGO~4.5 is available via GitHub\footnote{\url{https://github.com/yago-naga/yago-4.5}}  under a Creative Commons Attribution license. 

\section{Conclusion}\label{sec:conclusion}

In this paper, we have presented a method to merge the Wikidata taxonomy with the taxonomy of Schema.org, giving rise to a new version of the YAGO knowledge base, YAGO~4.5. Our work adds a rich layer of informative classes to YAGO, while at the same time keeping YAGO logically consistent. Going beyond YAGO, we have introduced, discussed, and justified several design principles that can be of use for other KB projects as well, most notably concerning the choice between properties vs classes, of properties vs inverse properties, and the modeling of fictional entities. We have also described implementation experiences that can help other researchers who work on  Wikidata.

Several open challenges remain: first, we did not include the classes of \emph{Bio\-Chem\-Entities} and \emph{Medical\-Entities}. These can be added in the future as a domain-specific extension. Now that the general framework is in place, more upper classes and properties can be defined. An interesting avenue of research in this direction would be the (automated) translation of textual class descriptions (such as \emph{de facto embassy} (Q5244910)) into logical properties and constraints~\cite{lu2022parsing}. This would require a better understanding of commonsense statements~\cite{nonnamed}. 
Another challenge is the maintenance: while the code can be rerun on newer versions of Wikidata, the manual mapping of classes might have to be adapted if the upper taxonomy of Wikidata changes -- as in previous versions of YAGO. Constraints, too, are currently defined manually. They could be mined automatically instead~\cite{amie,rabbani2023extraction,ortona2018robust,meilicke2019introduction}. 
The automated maintenance of structured data and its schema is an interesting challenge for the research community as a whole.

\begin{acks}
This work was partially funded by the \grantsponsor{ANR}{French National Agency for Research}{https://anr.fr}
under the Grant ``NoRDF'' \grantnum[https://nordf.telecom-paris.fr/en/]{ANR}{ANR-20-CHIA-0012-01}. 
\end{acks}

\bibliographystyle{splncs04}
\balance
\bibliography{bibliography}
\end{document}